\documentclass[11pt,letterpaper]{article}

\usepackage[margin=1in]{geometry}
\usepackage[T1]{fontenc}
\usepackage[utf8]{inputenc}
\usepackage{mathptmx}
\usepackage{courier}
\usepackage{amsmath,amssymb,amsthm}
\usepackage{graphicx,booktabs,array,microtype,flafter}
\usepackage[numbers,sort&compress]{natbib}
\usepackage[hidelinks]{hyperref}

\hypersetup{
  pdftitle={Schedule-Calibrated Commitment Order in Blockwise Masked Diffusion Language Models},
  pdfauthor={Gaurav Sarkar, Umang Pandey, Pratik Prabhanjan Brahma, Zicheng Liu, Emad Barsoum}
}
\newtheorem{proposition}{Proposition}
\newtheorem{corollary}{Corollary}
\title{Schedule-Calibrated Commitment Order in Blockwise Masked Diffusion Language Models}
\author{\normalsize Gaurav Sarkar\textsuperscript{1}\thanks{Work done while at AMD.}
\quad Umang Pandey\textsuperscript{2}
\quad Pratik Prabhanjan Brahma\textsuperscript{2}\\[3pt]
\normalsize Zicheng Liu\textsuperscript{2} \quad Emad Barsoum\textsuperscript{2}\\[6pt]
\small \textsuperscript{1}International Institute of Information Technology, Hyderabad\\[2pt]
\small \textsuperscript{2}Advanced Micro Devices (AMD)}
\date{}

\begin{document}
\maketitle

\begin{abstract}
Sequential block schedules can make token commitment in masked diffusion language models appear strongly left-to-right even when within-block order is random or reversed. For equal contiguous blocks processed without interleaving in positional order, with untied irreversible commitments, we use a Spearman-rank identity to express global position--commitment correlation as a deterministic schedule reference plus a weighted mean within-block correlation. In a reported paired study of LLaDA-8B-Instruct on 500 math and code prompts with 256 generated positions and updates, increasing block size from 64 to 128 lowers mean global correlation from 0.9845 to 0.9149. Uniform-random within-block references are already 0.9375 and 0.7500. After calibration, mean within-block correlations are 0.751 and 0.660, so the larger raw residual at block size 128 does not indicate stronger local positional ordering. We give sharp bounds and derive calibrated intervals from reported summaries, while documenting the limitations of the aggregate-only empirical release. The analysis supports schedule-aware interpretation of commitment order; its empirical scope is one checkpoint and two schedules, without evaluation of output quality or inference speed.
\end{abstract}

\section{Introduction}
\label{sec:introduction}

Masked diffusion language models predict missing tokens using a partially observed sequence. Their potential for parallel generation has motivated both blockwise inference and confidence-aware token selection~\citep{nie2025llada,arriola2025block,wu2025fast}. A position--commitment plot can nevertheless display a pronounced left-to-right wave. Interpreting that wave requires asking which orderings the model could actually have chosen: if an earlier block must finish before a later block can reveal any token, much of the global order is fixed by the schedule.

Consider 256 generated positions partitioned into four sequential blocks of size 64. Even if every block is revealed in exactly reverse positional order, the global position--commitment Spearman correlation remains approximately 0.875. Under uniformly random order within each block, its expectation is approximately 0.938. Thus, a high global correlation alone does not establish a strong left-to-right preference within the decisions available to the decoder. These values are analytical consequences of the schedule, not additional model experiments.

This paper gives a compact calibration for that measurement problem. Under sequential equal-sized blocks and untied commitments, a classical Spearman identity expresses the global correlation as a schedule reference plus a weighted mean of within-block correlations. The weight matters: subtracting the reference produces a residual whose scale still depends on block size. Dividing by that weight recovers the mean within-block correlation itself.

We apply this accounting to the aggregate results of a paired 500-prompt LLaDA-8B-Instruct study at block sizes 64 and 128. The global correlation decreases from 0.9845 to 0.9149, whereas the raw schedule-adjusted residual increases. Calibration resolves this apparent tension: mean within-block correlations are approximately 0.751 and 0.660. The residual increase is therefore not evidence of stronger local positional order at the larger block size.

Our contribution is a schedule-aware interpretation of a standard rank statistic. We provide (i) the exact reference, weight, and sharp bounds; (ii) a transparent distinction between reported empirical summaries and quantities derived from them; and (iii) a reporting protocol that avoids attributing schedule-constrained behavior to a model-intrinsic mechanism. This calibration describes commitment order rather than selecting an optimal decoding schedule.

\section{Related Work}
\label{sec:related}

\paragraph{Masked diffusion and blockwise inference.}
LLaDA supplies the masked-diffusion model family used in our study~\citep{nie2025llada}. Block Diffusion combines autoregressive organization across blocks with diffusion within blocks and includes a training framework~\citep{arriola2025block}. This is distinct from imposing a blockwise inference schedule on a fixed pretrained LLaDA checkpoint. Fast-dLLM uses approximate blockwise caching and confidence-aware parallel decoding~\citep{wu2025fast}. These methods motivate careful treatment of the schedule and selection rule as parts of the system being measured. Our untied, single-token analysis does not directly cover the simultaneous commitment events of parallel decoding.

\paragraph{Adaptive boundaries and deferred commitment.}
AdaBlock-dLLM adapts block boundaries using semantic-delimiter confidence~\citep{lu2026adablock}. Deferred Commitment Decoding maintains a certainty-aware moving window that defers uncertain positions~\citep{shu2026deferred}. GeoBlock selects block granularity using attention-derived dependency structure~\citep{wan2026geoblock}. These works already motivate why rigid boundaries can affect inference. Our objective is narrower: to calibrate an order statistic under a specified fixed-block process. The identity below is not an alternative scheduler, and its equal-block form does not automatically extend to those methods.

\paragraph{Interpretability of decoding order.}
DLM-Scope uses sparse autoencoders to examine feature stability before token commitment and feature drift afterward under different decoding orders~\citep{wang2026dlmscope}. Our analysis asks a complementary question that does not require a learned feature dictionary: how much global positional correlation is implied by block precedence before a representation-level explanation is considered?

\section{Commitment Order and Its Schedule Reference}
\label{sec:calibration}

\subsection{Definitions and assumptions}

For prompt $q$, let $N>1$ denote the number of generated positions. A schedule partitions them into $K=N/\sigma$ contiguous blocks of equal size $\sigma$, where $\sigma$ divides $N$. Blocks are processed sequentially in increasing position order: no position in block $k+1$ can commit until every position in block $k$ has committed. A commitment is the irreversible replacement of a masked position by a token; it is not the first time a provisional prediction matches the eventual token. A hidden representation may continue to change after its token is committed.

Let $t_j(q,\sigma)$ be the commitment update of position $j$. We assume that commitment times have no ties, as ensured by revealing exactly one previously masked position at each update. Their ranks $R_j$ form a permutation of $1,\ldots,N$. The primary per-prompt statistic is
\begin{equation}
\rho_\sigma(q)=\operatorname{corr}_{\mathrm{rank}}\bigl(j,t_j(q,\sigma)\bigr)
=\operatorname{corr}\bigl(j,R_j(q,\sigma)\bigr).
\label{eq:global}
\end{equation}
Within block $k$, let $\pi_k(i)$ be the local reveal rank of local position $i$, and define $r_k=\operatorname{corr}(i,\pi_k(i))$ for $\sigma>1$. Write $\bar r_\sigma=K^{-1}\sum_{k=1}^K r_k$ for the mean within-block correlation of one prompt.

\subsection{Exact decomposition and attainable range}

\begin{proposition}[Schedule-calibrated rank identity]
\label{prop:identity}
Under the assumptions above, for $1<\sigma\leq N$,
\begin{equation}
\boxed{\rho_\sigma=B_\sigma+W_\sigma\bar r_\sigma},\qquad
B_\sigma=\frac{N^2-\sigma^2}{N^2-1},\qquad
W_\sigma=\frac{\sigma^2-1}{N^2-1}.
\label{eq:identity}
\end{equation}
Uniformly random within-block permutations satisfy $\mathbb{E}[\bar r_\sigma]=0$, so $\mathbb{E}[\rho_\sigma]=B_\sigma$. Independence between blocks is not required for this expectation.
\end{proposition}

\begin{proof}[Proof sketch]
Sequential block completion implies $j=(k-1)\sigma+i$ and $R_j=(k-1)\sigma+\pi_k(i)$. The block offsets cancel in $j-R_j$. Substituting the within-block squared rank differences into the untied Spearman formula gives Equation~\eqref{eq:identity}. Appendix~\ref{app:proof} gives the full derivation.
\end{proof}

The identity holds for every admissible set of permutations, without a probabilistic assumption about model logits. Random within-block order supplies an interpretable reference, not a model of the actual decoding policy. In particular, $B_\sigma$ is not a lower bound.

\begin{corollary}[Attainable global correlations]
\label{cor:bounds}
The sharp bounds under the stated schedule are
\begin{equation}
\frac{N^2-2\sigma^2+1}{N^2-1}=B_\sigma-W_\sigma
\;\leq\;\rho_\sigma\;\leq\;B_\sigma+W_\sigma=1.
\label{eq:bounds}
\end{equation}
Reversing every block attains the lower bound; revealing every block left-to-right attains the upper bound.
\end{corollary}

For $N=256$, the lower bounds are approximately 0.8750 at $\sigma=64$ and 0.5000 at $\sigma=128$. A strongly positive global statistic can therefore coexist with perfectly negative local order. These bounds also show why the reference must be reported with the statistic. At $\sigma=1$, the schedule forces $\rho_1=1$, while a within-block correlation is undefined. At $\sigma=N$, $B_N=0$ and $W_N=1$, so global and within-block correlations coincide.

\subsection{Residuals, normalization, and interpretation}

Define the raw residual and calibrated statistic by
\begin{equation}
\delta_\sigma=\rho_\sigma-B_\sigma,
\qquad
C_\sigma=\frac{\rho_\sigma-B_\sigma}{W_\sigma}=\bar r_\sigma.
\label{eq:calibrated}
\end{equation}
Both may be negative. The residual is bounded by $[-W_\sigma,W_\sigma]$ and the calibrated statistic by $[-1,1]$, with attainable endpoints in both cases. Comparing residual magnitudes across block sizes without accounting for $W_\sigma$ can consequently reverse the interpretation of local ordering.

Calibration does not recover information absent from the within-block ranks. If those ranks are available, directly computing $\bar r_\sigma$ is equally valid. The decomposition instead connects local order to a commonly used global statistic, makes its schedule-dependent range explicit, and permits local-order summaries to be recovered from aggregate global summaries when $N$ and $\sigma$ are fixed.

A positive $C_\sigma$ describes positional order produced jointly by the checkpoint, prompt, confidence-selection rule, and schedule-conditioned visible context. It does not isolate a model-intrinsic tendency. Changing block size changes which tokens can become visible before other decisions; the residual is not the outcome of a controlled intervention on a hidden-state mechanism. Likewise, $B_\sigma/\rho_\sigma$ is not a fraction of explained variance or a causal attribution percentage.

\section{Paired Study: Reported Results and Calibration}
\label{sec:study}

\subsection{Protocol and provenance}

The empirical component uses the reported aggregate results of a paired study with 500 prompts, retained from the \href{https://openreview.net/forum?id=dJTCPQZINw}{earlier MRD manuscript (OpenReview, 2026)}. The sample contains 125 prompts each from GSM8K, MATH, MBPP, and HumanEval~\citep{cobbe2021gsm8k,hendrycks2021math,austin2021mbpp,chen2021humaneval}. Each prompt was evaluated with the same LLaDA-8B-Instruct checkpoint at $\sigma=64$ and $\sigma=128$, with generated length and total update budget fixed at $N=T=256$. In this design, each block receives $\sigma$ updates and each update reveals one token permanently.

The reported configuration uses \texttt{GSAI-ML/LLaDA-8B-Instruct}, its chat template, temperature zero, no EOS logit bias, selection seed 271828, and decode seed 1729. The reported checkpoint revision is \texttt{08b83a6feb34df1a6011b80c3c00c7563e963b07}. Under the sampler's \texttt{low\_confidence} selection convention, the highest-confidence eligible prediction is revealed and lower-confidence positions remain masked. The name does not mean that the least confident token is revealed first.

Correlations were computed separately for each prompt and then averaged. The reported intervals use prompt-level bootstrapping; the difference between schedules uses paired prompts. Tokens within a prompt are not treated as independent replicates. These experimental summaries and settings are retained from the original study, not independently rerun in this revision. The present release does not provide the exact dataset revisions and splits, selected prompt identifiers, or the original bootstrap configuration and records, including replicate count, random seed, and interval construction.

Let $Q=500$ and $\widehat\mu_\rho(\sigma)=Q^{-1}\sum_q\rho_\sigma(q)$. Because $B_\sigma$ and $W_\sigma$ are fixed across these prompts, Equation~\eqref{eq:identity} also holds for the means:
\begin{equation}
\widehat\mu_r(\sigma)=\frac{\widehat\mu_\rho(\sigma)-B_\sigma}{W_\sigma}.
\label{eq:means}
\end{equation}
The residual and calibrated intervals below are positive affine transformations of the reported global interval endpoints. They are not a new bootstrap or a reanalysis of raw model outputs.

\subsection{Global order and within-block order}

\begin{figure}[tbp]
\centering
\includegraphics[width=0.84\linewidth]{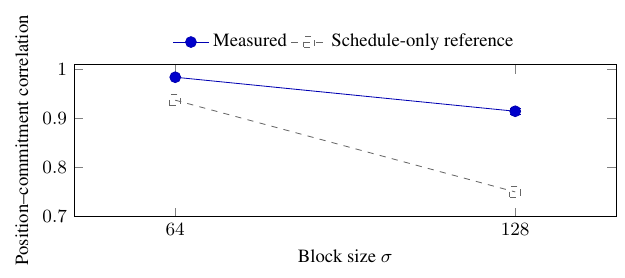}
\caption{Reported mean global commitment correlations for the same 500 prompts and the analytic uniform-random within-block references. Error bars reproduce the reported 95\% prompt-bootstrap intervals; the interval at $\sigma=64$ is smaller than the marker. Lines connect only the two evaluated schedules and do not imply a measured trend at intermediate block sizes.}
\label{fig:commitment}
\end{figure}

\begin{table}[tbp]
\centering
\caption{Global summaries retained from the original study and algebraically calibrated quantities. Brackets give the reported 95\% global interval and its affine transformations for the residual and local correlation. The reference and weight are deterministic, not estimated. Transformed intervals use the exact constants and rounded reported endpoints; no new resampling is performed.}
\label{tab:primary}
\small
\setlength{\tabcolsep}{4pt}
\begin{tabular}{@{}lcc@{}}
\toprule
Quantity & $\sigma=64$ & $\sigma=128$ \\
\midrule
Global mean $\widehat\mu_\rho$ & 0.984457 & 0.914898 \\
Reported interval & [0.983510, 0.985360] & [0.908403, 0.921276] \\
Schedule reference $B_\sigma$ & 0.937514 & 0.750011 \\
Within-block weight $W_\sigma$ & 0.062486 & 0.249989 \\
\addlinespace
Derived mean residual & 0.046943 & 0.164887 \\
Transformed interval & [0.045996, 0.047846] & [0.158392, 0.171265] \\
Derived local mean $\widehat\mu_r$ & 0.751 & 0.660 \\
Transformed interval & [0.736, 0.766] & [0.634, 0.685] \\
\bottomrule
\end{tabular}
\end{table}

Table~\ref{tab:primary} and Figure~\ref{fig:commitment} show that the global mean decreases from 0.984457 to 0.914898. The reported paired global difference is
\begin{equation}
\widehat\mu_\rho(64)-\widehat\mu_\rho(128)
=0.069559,\qquad 95\%\text{ interval }[0.063599,0.075711].
\label{eq:paired}
\end{equation}
The global correlation decreases for 97.4\% of the paired prompts. This is a statement about global $\rho$, not the proportion of prompts with decreasing calibrated local order.

Both schedule-adjusted mean residuals are positive, with transformed intervals above zero. Under the reported uncertainty estimates, the decoder exhibits positive within-block positional order beyond the uniform-random reference. However, the residual at $\sigma=128$ is larger partly because its weight is approximately four times the weight at $\sigma=64$. Normalizing gives mean local correlations of 0.751 and 0.660. This distinction is the empirical illustration of why a raw residual is not a block-size-independent measure.

The paired interval in Equation~\eqref{eq:paired} cannot be reused as an interval for $\widehat\mu_r(64)-\widehat\mu_r(128)$: the two schedules have different weights. Recovering the uncertainty of that paired local-order contrast requires the per-prompt data or the corresponding joint bootstrap replicates. Nor is decreasing global correlation with increasing block size a theorem. Different local orders can offset, or reverse, the change in the schedule reference.

\subsection{Terminal-token sensitivity}

\begin{table}[tbp]
\centering
\caption{Reported terminal-token sensitivity summaries. Content-only correlations exclude terminal tokens and recompute ranks. They are not calibrated with the original equal-block formula.}
\label{tab:terminal}
\small
\begin{tabular}{@{}lcc@{}}
\toprule
Quantity & $\sigma=64$ & $\sigma=128$ \\
\midrule
Content-only mean $\rho$ & 0.989410 & 0.941235 \\
Reported 95\% interval & [0.988666, 0.990131] & [0.935613, 0.946779] \\
EOS fraction & 0.051664 & 0.053414 \\
\bottomrule
\end{tabular}
\end{table}

The content-only means retain the same ordering, while the reported EOS fractions are similar (Table~\ref{tab:terminal}). This check is consistent with the pooled global difference not arising solely from the included terminal positions. It does not replace a full token-level audit. Removing terminal tokens can leave unequal retained counts in different blocks; consequently, substituting a shorter length into Equation~\eqref{eq:identity} would not in general yield the correct calibration. We report this sensitivity check without a schedule-adjusted residual.

\section{Implications, Limitations, and Availability}
\label{sec:limitations}

\paragraph{What the diagnostic supports.}
A commitment-order report should state the scheduling constraints before interpreting its global correlation. In the setting analyzed here, reporting $N$, $\sigma$, the tie policy, $B_\sigma$, $W_\sigma$, and $\bar r_\sigma$ makes the observation interpretable. A diagonal position--time plot or a high global correlation is insufficient on its own to infer learned autoregressive behavior. Calibration is useful for that interpretation; it neither removes every confound nor certifies a good schedule.

\paragraph{Empirical scope.}
The study covers one checkpoint, two block sizes, one generated length, and one update budget. Its pooled summaries cannot establish that the same effect holds within each of the four datasets. No task-answer accuracy, executable-code score, quality--latency trade-off, or wall-clock acceleration is evaluated here. Equal update counts do not by themselves guarantee equal runtime or computational cost. The commitment measurements also provide no validated early-stopping criterion: a token becoming fixed does not imply that a complete answer is correct.

\paragraph{Schedules outside the assumptions.}
Adaptive or unequal blocks, a shorter final block, overlapping windows, tied parallel commitments, and remasking require separate treatment. In particular, untied Spearman's squared-difference formula cannot simply be reused for average ranks with ties. Our results should not be read as tests of all blockwise or pure-diffusion decoders. The full-block case $\sigma=N$ is covered by the identity, but the present empirical comparison does not include it.

\paragraph{Representation-level explanations remain open.}
The rank identity is agnostic to hidden representations. It does not establish prompt-conditioned manifold routing, representation collapse, or invariance of hidden dynamics across schedules. Connecting rank order to geometry requires separately specified measurements and controlled or held-out tests linked to token-level decisions. Geometry summaries are not used as empirical support for the commitment-order result.

\paragraph{Data and code availability.}
The source release contains the reported aggregate summaries and a standard-library Python utility that checks the rank identity on synthetic permutations and reproduces the affine calculations in Table~\ref{tab:primary}. These are algebra and arithmetic checks, not replication of the 500-prompt model study. The release does not include the 1,000 per-prompt reveal-rank sequences, terminal-token flags, selected-prompt manifest, generated completions, or original bootstrap outputs. Accordingly, the reported empirical intervals and sensitivity results cannot be independently recomputed from the released artifacts. Appendix~\ref{app:reproducibility} specifies the records needed to close this gap.

\section{Conclusion}

Sequential block precedence can force a high global commitment-order correlation even when local order is random or reversed. An elementary rank decomposition provides its exact reference and scale, while the normalized residual equals mean within-block order. Applied to the reported paired study, it distinguishes a decreasing global correlation from an increasing raw residual and yields positive local-order means for both evaluated schedules. The appropriate conclusion is a reporting principle: interpret commitment order relative to the decisions the schedule permits. General claims about model mechanisms, output quality, and inference efficiency require additional evidence.

\bibliographystyle{unsrtnat}
\bibliography{references}

@misc{nie2025llada,
  author = {Shen Nie and Fengqi Zhu and Zebin You and Xiaolu Zhang and Jingyang Ou and Jun Hu and Jun Zhou and Yankai Lin and Ji-Rong Wen and Chongxuan Li},
  title = {Large Language Diffusion Models},
  howpublished = {arXiv:2502.09992},
  year = {2025},
  url = {https://arxiv.org/abs/2502.09992}
}

@inproceedings{arriola2025block,
  author = {Marianne Arriola and Aaron Gokaslan and Justin T. Chiu and Zhihan Yang and Zhixuan Qi and Jiaqi Han and Subham Sekhar Sahoo and Volodymyr Kuleshov},
  title = {Block Diffusion: Interpolating Between Autoregressive and Diffusion Language Models},
  booktitle = {International Conference on Learning Representations},
  year = {2025},
  note = {arXiv:2503.09573},
  url = {https://arxiv.org/abs/2503.09573}
}

@misc{wu2025fast,
  author = {Chengyue Wu and Hao Zhang and Shuchen Xue and Zhijian Liu and Shizhe Diao and Ligeng Zhu and Ping Luo and Song Han and Enze Xie},
  title = {{Fast-dLLM}: Training-free Acceleration of Diffusion {LLM} by Enabling {KV} Cache and Parallel Decoding},
  howpublished = {arXiv:2505.22618},
  year = {2025},
  url = {https://arxiv.org/abs/2505.22618}
}

@inproceedings{lu2026adablock,
  author = {Guanxi Lu and Hao Mark Chen and Yuto Karashima and Zhican Wang and Daichi Fujiki and Hongxiang Fan},
  title = {{AdaBlock-dLLM}: Semantic-Aware Diffusion {LLM} Inference via Adaptive Block Size},
  booktitle = {International Conference on Learning Representations},
  year = {2026},
  note = {arXiv:2509.26432},
  url = {https://arxiv.org/abs/2509.26432}
}

@misc{shu2026deferred,
  author = {Yingte Shu and Yuchuan Tian and Chao Xu and Yunhe Wang and Hanting Chen},
  title = {Deferred Commitment Decoding for Diffusion Language Models},
  howpublished = {arXiv:2601.02076},
  year = {2026},
  url = {https://arxiv.org/abs/2601.02076}
}

@misc{wan2026geoblock,
  author = {Lipeng Wan and Junjie Ma and Jianhui Gu and Zeyang Liu and Xuyang Lu and Xuguang Lan},
  title = {{GeoBlock}: Inferring Block Granularity from Dependency Geometry in Diffusion Language Models},
  howpublished = {arXiv:2603.26675},
  year = {2026},
  url = {https://arxiv.org/abs/2603.26675}
}

@misc{wang2026dlmscope,
  author = {Xu Wang and Bingqing Jiang and Yu Wan and Baosong Yang and Lingpeng Kong and Difan Zou},
  title = {{DLM-Scope}: Mechanistic Interpretability of Diffusion Language Models via Sparse Autoencoders},
  howpublished = {arXiv:2602.05859},
  year = {2026},
  url = {https://arxiv.org/abs/2602.05859}
}

@misc{cobbe2021gsm8k,
  author = {Karl Cobbe and Vineet Kosaraju and Mohammad Bavarian and Mark Chen and Heewoo Jun and Lukasz Kaiser and Matthias Plappert and Jerry Tworek and Jacob Hilton and Reiichiro Nakano and Christopher Hesse and John Schulman},
  title = {Training Verifiers to Solve Math Word Problems},
  year = {2021},
  howpublished = {arXiv:2110.14168},
  url = {https://arxiv.org/abs/2110.14168}
}

@misc{hendrycks2021math,
  author = {Dan Hendrycks and Collin Burns and Saurav Kadavath and Akul Arora and Steven Basart and Eric Tang and Dawn Song and Jacob Steinhardt},
  title = {Measuring Mathematical Problem Solving With the {MATH} Dataset},
  year = {2021},
  howpublished = {arXiv:2103.03874},
  url = {https://arxiv.org/abs/2103.03874}
}

@misc{austin2021mbpp,
  author = {Jacob Austin and Augustus Odena and Maxwell Nye and Maarten Bosma and Henryk Michalewski and David Dohan and Ellen Jiang and Carrie Cai and Michael Terry and Quoc Le and Charles Sutton},
  title = {Program Synthesis with Large Language Models},
  year = {2021},
  howpublished = {arXiv:2108.07732},
  url = {https://arxiv.org/abs/2108.07732}
}

@misc{chen2021humaneval,
  author = {Mark Chen and Jerry Tworek and Heewoo Jun and Qiming Yuan and Henrique Ponde de Oliveira Pinto and Jared Kaplan and Harri Edwards and Yuri Burda and Nicholas Joseph and Greg Brockman and Alex Ray and Raul Puri and Gretchen Krueger and Michael Petrov and Heidy Khlaaf and Girish Sastry and Pamela Mishkin and Brooke Chan and Scott Gray and Nick Ryder and Mikhail Pavlov and Alethea Power and Lukasz Kaiser and Mohammad Bavarian and Clemens Winter and Philippe Tillet and Felipe Petroski Such and Dave Cummings and Matthias Plappert and Fotios Chantzis and Elizabeth Barnes and Ariel Herbert-Voss and William Hebgen Guss and Alex Nichol and Alex Paino and Nikolas Tezak and Jie Tang and Igor Babuschkin and Suchir Balaji and Shantanu Jain and William Saunders and Christopher Hesse and Andrew N. Carr and Jan Leike and Josh Achiam and Vedant Misra and Evan Morikawa and Alec Radford and Matthew Knight and Miles Brundage and Mira Murati and Katie Mayer and Peter Welinder and Bob McGrew and Dario Amodei and Sam McCandlish and Ilya Sutskever and Wojciech Zaremba},
  title = {Evaluating Large Language Models Trained on Code},
  year = {2021},
  howpublished = {arXiv:2107.03374},
  url = {https://arxiv.org/abs/2107.03374}
}

\appendix
\section{Full Derivation and Boundary Cases}
\label{app:proof}

For untied ranks $j,R_j\in\{1,\ldots,N\}$, Spearman's identity gives
\begin{equation}
\rho_\sigma=1-\frac{6\sum_{j=1}^N(j-R_j)^2}{N(N^2-1)}.
\label{eq:spearman}
\end{equation}
For a sequential equal-block schedule, write $j=(k-1)\sigma+i$ and $R_j=(k-1)\sigma+\pi_k(i)$. Hence
\begin{equation}
\rho_\sigma=1-\frac{6\sum_{k=1}^K\sum_{i=1}^\sigma(i-\pi_k(i))^2}{N(N^2-1)}.
\end{equation}
Within block $k$,
\begin{equation}
r_k=1-\frac{6\sum_{i=1}^\sigma(i-\pi_k(i))^2}{\sigma(\sigma^2-1)},
\qquad
\sum_{i=1}^\sigma(i-\pi_k(i))^2=\frac{\sigma(\sigma^2-1)}{6}(1-r_k).
\end{equation}
Substitution, together with $K\sigma=N$, gives
\begin{align}
\rho_\sigma
&=1-\frac{\sigma(\sigma^2-1)\sum_{k=1}^K(1-r_k)}{N(N^2-1)}\\
&=1-\frac{\sigma^2-1}{N^2-1}(1-\bar r_\sigma)
=\frac{N^2-\sigma^2}{N^2-1}+\frac{\sigma^2-1}{N^2-1}\bar r_\sigma.
\end{align}
This is a deterministic equality, not an asymptotic approximation. Since each $r_k\in[-1,1]$, Equation~\eqref{eq:bounds} follows immediately. The permutations $\pi_k(i)=\sigma+1-i$ and $\pi_k(i)=i$ attain its endpoints.

For a uniformly random permutation, $\mathbb{E}[\pi_k(i)]=(\sigma+1)/2$. The expected centered-rank covariance is therefore zero, whereas its normalization is fixed. Thus $\mathbb{E}[r_k]=0$ and linearity gives $\mathbb{E}[\rho_\sigma]=B_\sigma$, without independence between blocks. At $\sigma=1$, all ranks are forced and the result is $\rho_1=1$ directly; one must not divide by $W_1=0$. At $\sigma=N$, the reference vanishes, but the model's correlation need not vanish.

The ancillary check computes correlations directly from centered ranks using exact rational arithmetic and compares them with the identity. It also checks reverse-order bounds. Such enumeration can detect implementation mistakes in the formula, but cannot validate any reported model-generated trajectory.

\section{Interval Transformations and Reproducibility Records}
\label{app:reproducibility}

\subsection{What is and is not recomputed}

Let $[L_\sigma,U_\sigma]$ denote the reported interval accompanying the estimate $\widehat\mu_\rho(\sigma)$. With fixed $N$ and $\sigma>1$, the corresponding affine interval transformations are
\begin{align}
I_\delta(\sigma)&=[L_\sigma-B_\sigma,\ U_\sigma-B_\sigma],\\
I_r(\sigma)&=\left[\frac{L_\sigma-B_\sigma}{W_\sigma},\ \frac{U_\sigma-B_\sigma}{W_\sigma}\right].
\end{align}
These transformations preserve the coverage statement of the input interval under the exact relationship; they do not validate that input interval or reveal how it was originally constructed. The values for $N=256$ are
\begin{equation}
B_{64}=\frac{4096}{4369},\quad W_{64}=\frac{273}{4369},\qquad
B_{128}=\frac{16384}{21845},\quad W_{128}=\frac{5461}{21845}.
\end{equation}
We apply these exact constants to the global means and endpoints reported to six decimal places. Residuals are displayed to six decimals and normalized correlations to three decimals; the latter transformation amplifies input rounding, especially at $\sigma=64$. The intervals are therefore approximate transformations of rounded reported endpoints, not recovered raw-data estimates. This revision uses these internally consistent transformed intervals in place of the separately reported residual intervals. The provenance of the earlier small interval discrepancy is not established by the available records.

The paired global interval does not determine a paired local-order interval. A paired local difference has the form
\begin{equation}
\frac{\rho_{64}-B_{64}}{W_{64}}-\frac{\rho_{128}-B_{128}}{W_{128}},
\end{equation}
which weights the two prompt-level correlations differently. Marginal intervals and an interval for the unweighted global difference do not determine its joint resampling distribution.

\subsection{Records required for an empirical reproduction}

A complete reproduction would need a prompt manifest containing dataset name, exact revision and split, example identifier, final formatted prompt or its reproducible construction, and selection settings. Each prompt--schedule record should contain $N$, the block boundaries, checkpoint and tokenizer revisions, decoding configuration, final token IDs, one reveal rank per position, and the terminal-token mask used in the sensitivity analysis. The rank vector must be a permutation of $1,\ldots,N$ under this design, and each block's ranks must occupy its prescribed contiguous range.

An analysis script could then compute $\rho_\sigma(q)$ from those records, compute all $r_k(q)$ independently, check the identity per record, and form prompt-level means. Paired resampling would use the same sampled prompt indices for both schedules. The release should specify whether resampling is pooled or stratified by dataset, its replicate count and seed, its interval method, and the terminal-token definition. These are requirements for a future reproducible analysis, not claims about settings recovered in this revision.

Per-dataset reporting and correctness scores would test whether the pooled association is consistent across tasks and whether it is related to useful outputs. Runtime claims would additionally require hardware and software versions, precision, batching, cache configuration, warm-up policy, and measured latency. These analyses remain outside the evidence supplied here.

\end{document}